%% file: latex/acl_latex.tex
\pdfoutput=1

\documentclass[11pt]{article}

\usepackage{hyperref}
\usepackage{url}
\usepackage{afterpage}
\usepackage{graphicx}
\usepackage{booktabs}
\usepackage{algorithm}
\usepackage{algorithmic}
\usepackage{booktabs}
\usepackage{placeins}
\usepackage{times}
\usepackage{latexsym}
\usepackage{cancel}
\usepackage{tikz}
\usepackage{tcolorbox}
\usepackage{supertabular}
\usepackage{longtable}

\makeatletter
\let\ST@trace\relax
\makeatother
\usepackage{xtab, booktabs}

\usepackage{amsmath,amsfonts,amssymb}

\usepackage{xcolor}         
\usepackage{amsmath}
\usepackage{soul}

\usepackage{subcaption}

\usepackage{cinzel} 
\usepackage{multirow}
\usepackage{multicol}
\usepackage{tabularx}
\usepackage{fontawesome5}

\usepackage{comment} 
\usepackage{wrapfig} 
\usepackage{bbm}
\usepackage{pifont}
\usepackage{makecell,rotating}
    \setcellgapes{5pt}
\usepackage{calligra}

\usepackage{aurical}
\usepackage{listings}
\usepackage{color}
\usepackage{xcolor}
\usepackage[T1]{fontenc}
\usepackage{numprint}

\usepackage[final]{acl}

\usepackage{times}
\usepackage{latexsym}

\usepackage[T1]{fontenc}

\usepackage[utf8]{inputenc}

\usepackage{microtype}

\usepackage{inconsolata}

\usepackage{graphicx}

\newcommand{\sysname}{\textsc{JOSH}}
\title{Sparse Rewards Can Self-Train Dialogue Agents}

\author{Barrett Martin Lattimer, Varun Gangal, Ryan McDonald, and Yi Yang \\
ASAPP\\
\texttt{blattimer@asapp.com} \\
}

\begin{document}
\maketitle

\begin{abstract}
\input{sections/abstract}
\end{abstract}

\begin{figure*}[t!]
    \centering
    \includegraphics[width=0.8\textwidth]{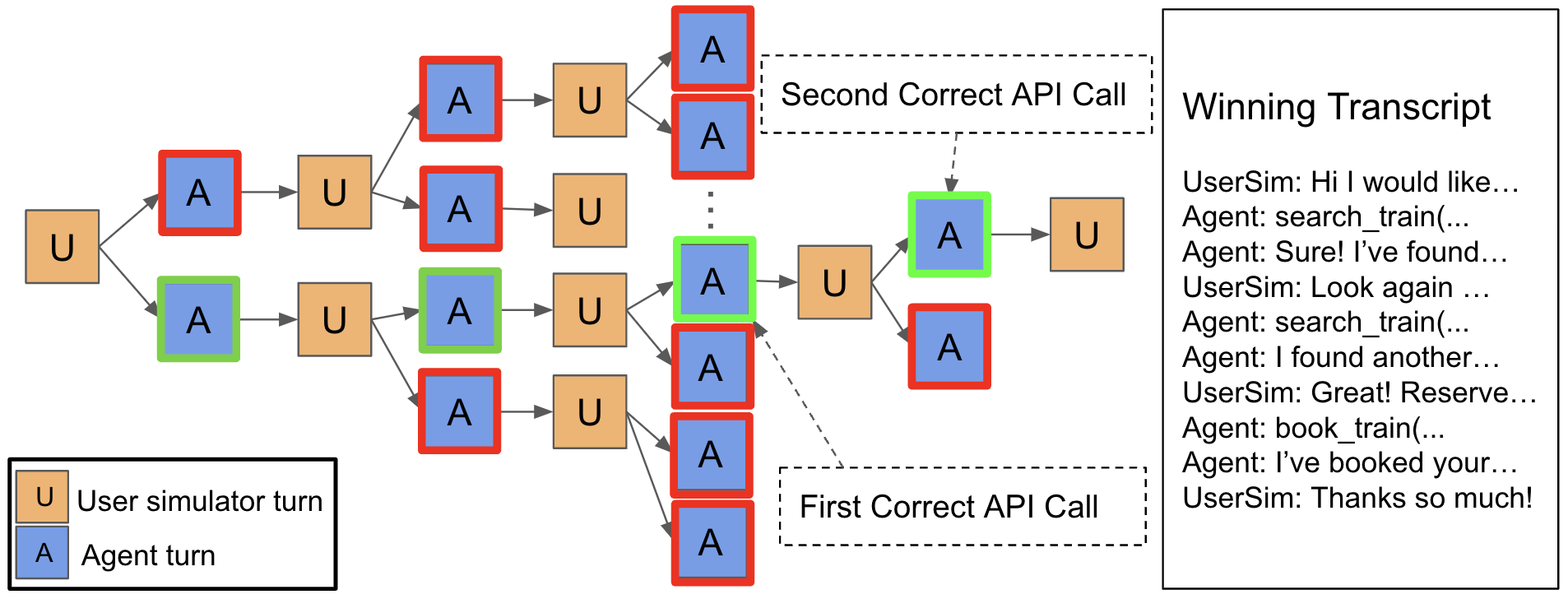}
    \caption{JOSH algorithm visualization: A simulation-based beam search traces optimal conversation paths (in green) by backtracking from successful API calls. This reveals ideal conversation flows and generates training data.} 
    \label{fig:beam_search}
\end{figure*}

\input{sections/introduction}

\input{sections/method}

\input{sections/data}

\input{sections/experiment}

\input{sections/results}

\input{sections/analysis}

\input{sections/related}

\input{sections/conclusion}

\input{sections/limitations}

\bibliography{acl2025_conference}

\clearpage
\appendix
\section{Appendix}
\input{sections/appendix}

\end{document}

%% file: sections/abstract.tex
Recent advancements in state-of-the-art (SOTA) Large Language Model (LLM) agents, especially in multi-turn dialogue tasks, have been primarily driven by supervised fine-tuning and high-quality human feedback. However, as base LLM models continue to improve, acquiring meaningful human feedback has become increasingly challenging and costly. In certain domains, base LLM agents may eventually exceed human capabilities, making traditional feedback-driven methods impractical. In this paper, we introduce a novel self-improvement paradigm that empowers LLM agents to autonomously enhance their performance without external human feedback. Our method, Juxtaposed Outcomes for Simulation Harvesting (JOSH), is a self-alignment algorithm that leverages a sparse reward simulation environment to extract ideal behaviors and further train the LLM on its own outputs. We present ToolWOZ, a sparse reward tool-calling simulation environment derived from MultiWOZ. We demonstrate that models trained with JOSH, both small and frontier, significantly improve tool-based interactions while preserving general model capabilities across diverse benchmarks. Our code and data are publicly available on GitHub at \url{https://github.com/asappresearch/josh-llm-simulation-training.git}

%% file: sections/introduction.tex
\section{Introduction}
\label{sec:intro}

Large Language Models (LLMs) \citep{bommasani2021opportunities, brown2020language, achiam2023gpt} have shown a well-marked ability to follow instructions under various tasks. These advancements are often attributed to post-training fine-tuning based on human preferences. This includes multi-turn tool calling tasks where an LLM-based agent must solve a task by interacting with both a user and a set of tools (or APIs) \citep{farn2023tooltalk,yao2024tau}. Further task-specific alignment for tool-calling tasks can take the form of preference judgments. But these can be expensive to obtain. Furthermore, there is usually a more `crisp' notion of success for such tasks. Specifically, was the right tool(s) or API(s) called with the right set of arguments? Ideally, alignment should be optimized towards these sparse rewards.

In this paper, we propose a self-alignment process \sysname~(Juxtaposed Outcomes for Simulation Harvesting) that can be used to improve a model's performance on multi-turn tool calling by optimizing for tool/API completion using simulated rollouts of reward-conditioned conversations. We show that this method is general and can be applied to weak/small or frontier LLMs, though gains are significantly larger for the former. We also present a new tool calling benchmark, ToolWOZ, refashioning MultiWoz2.0 \citep{zang2020multiwoz} to train and evaluate multi-turn tool calling effectively.

\sysname~utilizes a beam search inspired simulation approach, employing sparse rewards (in this paper corresponding to successful tool calls) to guide turn-by-turn generation and synthesize preference-annotated examples. \sysname~allows an agent to generate multiple responses at each conversation turn, exploring various trajectories through a conversation until a sparse reward (goal) is encountered along some path. Upon reaching a goal, other beam candidates are pruned and further expansion proceeds only from that point. Once a trajectory achieves all possible goals, all remaining trajectories are backtracked, logging unsuccessful paths as negative alignment samples and successful paths as positive ones. This process constructs alignment preference data solely from the model itself. When used to align that same model, we show it enhances model performance.

Tool use is a critical skill in LLMs \citep{mialon2023augmented,schick2024toolformer}; however, there is a large disparity in tool-using capabilities across different model sizes, especially when involving them in multi-turn reasoning \citep{gao2024efficient}. Furthermore, existing benchmarks either lack a real-world multi-turn setup \citep{ruan2024toolemu} or intentionally keep the agent's dialogue disjoint from underlying databases and focus more on tool selection \citep{huangmetatool}. To demonstrate \sysname's capability to improve a model used in an agentic system through self-alignment, we introduce a new dataset ToolWOZ. Derived from the task-oriented dialogue (TOD) benchmark MultiWOZ, ToolWOZ is a multi-turn tool-based simulation environment where an agent model is assessed on its tool-calling capabilities by calling goal APIs through collaboration with a user simulator. After self-alignment using \sysname~on the ToolWOZ training set, a \texttt{meta-llama3-8B-instruct} \citep{Llama38BMeta} model exhibits a 74\% increase in Success Rate and outperforms PPO \cite{schulman2017proximal}. Additionally, after JOSH self-alignment we see \texttt{gpt-4o} beats it's own baseline to become state-of-the-art on two separate benchmarks: ToolWOZ and $\tau$-bench \citep{yao2024tau}.

To show that \sysname~does not degrade general model performance, we evaluate a trained \texttt{meta-llama3-8B-instruct} model across three public benchmarks: MT-Bench \citep{zheng2024judging}, $\tau$-bench , and the LMSYS-Chatbot Arena Human Preference Predictions challenge \citep{lmsys-chatbot-arena, zheng2024judging}. Our results confirm that the \sysname~aligned model does not regress relative to its base model, despite its new specialized alignment.

%% file: sections/method.tex
\section{JOSH: Juxtaposed Outcomes for Simulation Harvesting}
\label{sec:method}

In this section, we detail the two components of JOSH, our method for generating synthetic preference-annotated training data to enhance tool-driven multi-turn dialogue. We use the terms ``tool'' and ``API'' interchangeably. Our approach for generating conversations uses an agent-simulator system and involves a turn-level beam search strategy combined with tool/API-calling reward pruning. Unlike traditional token-level beam search, our method maintains multiple active multi-turn conversations (trajectories) over sequences of agent turns. From these synthetic conversations we create preference-annotated training instances. This involves extracting both supervised fine-tuning (SFT) data and preference fine-tuning (PFT) data. The SFT data is derived from the optimal trajectory (or path) through the conversation tree, while the PFT data includes pairwise comparisons of good and bad agent turns, facilitating more nuanced model training.

\subsection{Beam Search Simulation for Collecting User-Agent Conversations}
\label{subsec:beamsearch_simulation}
{\begin{algorithm}[t]
\scriptsize
\caption{Beam Search Simulation for Collecting User-Agent Conversations}
\label{alg:beam}
\begin{algorithmic}[1]
\STATE Input parameters: max\_depth; branching\_factor; max\_beam
\STATE Load: $U \leftarrow \text{User Simulator}$; $A \leftarrow \text{Agent}$; $G \leftarrow \text{Goal Set}$\\
$\quad *$ $U(l)$ and $A(l)$ run one turn of the user and agent on a leaf node $l$ of a conversation trajectory.\\\ \\
\STATE // Initialize control parameters and data structures
\STATE $AR \leftarrow 0 $\ \ \ \ \ \ \ \ \ // Average reward
\STATE $leaves \leftarrow []$\ \ // Trajectory leaf nodes which will be expanded in beam search
\STATE $depth \leftarrow 0$\ \ \ \ // Current trajectory depth \\\ \\

\WHILE{depth $\leq$ max\_depth \AND $G\neq \emptyset$}
  \STATE // Expand trajectories by running user simulation and agent.
  \STATE $leaves = [U(l)\;\; \forall l \in leaves]$\ \ \ // Expand with next user response by running the simulator one step on all leaves.
  \STATE // Expand trajectories by running agent $A$ on all leaves.
  \STATE // Each expansion is a full turn of $A$ including API calls, thoughts  and utterances.
  \IF{$\text{len}(leaves) * \textit{branching\_factor} \leq \textit{max\_beam}$}
      \STATE $leaves = [A(l)\;\; \forall l \in leaves]$
  \ENDIF\\\ \\
  \STATE // Check for goals. If a goal is reached, prune trajectories to retain successful path.
  \FOR{$\textit{leaf} \in leaves$}
    \STATE // If a goal was reached in leaf
    \IF{$\exists g \in G$ \AND $g \in \textit{leaf}]$}
        \STATE // Set leaf as the new root, remove $g$ from Goal Set and update the reward.
        \STATE $leaves = [\textit{leaf}]$
        \STATE $G.remove(g)$
        \STATE $AR=AR+\frac{1}{\textit{len}(\text{Goal Set})}$
        \STATE \textbf{BREAK}
    \ENDIF
  \ENDFOR\\\ \\
  \STATE $depth \leftarrow depth + 1$
\ENDWHILE
\end{algorithmic}
\end{algorithm}}
We begin by having the agent \( A \) (defined in Section \ref{subsec:agent_methods}) interact with a user simulator \( U \) (defined in Section \ref{subsec:user_simulator_methods}). 
A set of goals \( G = \{g_1, g_2, \ldots, g_k\} \) is defined where achieving a goal will award the agent \( A \) a sparse reward of value $\frac{1}{\textit{len}(\text{Goal Set})}$ to it's return. Our return uses the Average Reward formulation \citep{sutton2018reinforcement} hence we denote it as \( AR \) and refer to it as ``Average Reward''. 

We considered several reward structures for the task of agent dialogues, we found that the cumulative reward method encourages excessive API calls, leading to inefficiency, which is contrary to our aim of minimal interaction for issue resolution. Per-turn rewards, while potentially speeding up learning, necessitate costly annotations or the use of a resource-intensive LLM judge, which we reserve for future exploration. Sparse goal-based rewards, akin to our approach, issue rewards only upon goal completion, offering feedback at each API call to refine agent behavior in real time. While shaped rewards might expedite learning by guiding agents with intermediate incentives, they complicate the reward structure and risk diverting focus from final objectives. By employing an average reward function with partial sparse rewards, we facilitate efficient task completion without the complexities of other structures, ensuring goal-oriented and concise dialogues.

We begin with  \( AR = 0 \). Goals in \( G \) can be achieved when \( A \) interacts with \( U \) in a desired manner. In this paper, rewards are granted when the agent successfully makes a predefined correct tool or API call during a conversation. Figure \ref{fig:beam_search} illustrates an example where the goal set  \( G \) is composed of multiple correct API calls made within a simulated conversation.

The beam search simulation, in which agent \( A \) and user simulator \( U \) interact, is detailed in Algorithm \ref{alg:beam}. The algorithm begins by constructing a tree: the user simulator \( U \) initiates the conversation, and agent \( A \) generates \textit{branching\_factor} agent responses with a high temperature to encourage variability. Each agent turn \( A(l) \) -- where $l$ is the leaf node of a conversation trajectory -- represents a full response, during which the agent may make API calls or other actions before replying to the user.

Following each agent turn, \( U \) generates a response, after which \( A \) generates another set of \textit{branching\_factor} turns for each response from \( U \). This continues until an agent turn achieves a goal \( g \). In the case of Figure \ref{fig:beam_search}, the goal is the ``First Correct API Call.'' The agent turn that achieves this goal becomes the new root, \( g \) is removed from the goal set \( G \), the Average Reward is increased by $\frac{1}{\textit{len}(\text{Goal Set})}$, and the process repeats. The goal \( g \) is removed from the goal set in order to prevent rewards for duplicate goals. If another turn simultaneously achieves the goal \( g \), it is considered partial credit: it does not follow the ideal path but is not considered a negative example either. When the number of branches reaches the \textit{max\_beam} size, only one agent response is generated. This process continues until all the goals in the goal set \( G \) are achieved or a maximum number of turns is reached. Because the beam search is designed to follow paths once goals are hit, this naturally selects for trajectories where goals are achieved earlier in the conversation. 

    

In this paper, we branch at the turn level rather than the agent action level, allowing the tree to grow exponentially with the number of turns rather than individual actions (i.e., utterance, thoughts, API/tools). Binary trees have a number of $2^{h-1}$ leaf nodes where h is the height of the tree, since JOSH splits at the turn level we can expect $t=\log_2(max\_beam)+1$ to be the number of turns $t$ before JOSH can no longer expand the tree. There are roughly 3 actions a per turn on average, so the number of branching turns allowed when when action splitting is $t=\frac{\log_2(max\_beam)+1}{3}$. Thus when $max\_beam=8$ which is used throughout the paper to keep costs reasonable (around \$100) we could perform either $t=4$ turns while turn splitting, or $t=\frac{4}{3}$ turns when splitting on actions. While splitting on actions may provide more diversity, over the course of a multi turn dialogue we can explore more possible paths deep in the tree for the same $max\_beam$ when splitting on turns.


\subsection{Preference Data Extraction}
\label{subsec:training_data_extract}
Once Algorithm \ref{alg:beam} terminates, we have a tree that resembles Figure \ref{fig:beam_search}, from which we can extract training data for both Supervised Fine-Tuning (SFT) and Preference Fine-Tuning (PFT).

For SFT, training data is created by backtracking up the tree from the final successful turn to the initial user-simulated utterance. We refer to this as the ``ideal path,'' illustrated by following the green agent turns up the tree in Figure \ref{fig:beam_search}, starting from the ``Second Correct API Call.'' This ideal path corresponds to the best agent turns generated to maximize the number of rewards achieved. This data can subsequently be used to train models, guiding them to produce responses that are likely to yield higher rewards. This approach is similar to offline RL with Decision Transformers, where an optimal path is found by conditioning on the reward \citep{chen2021decision}.

For PFT, we use the same tree but additionally take advantage of suboptimal model outputs. We create pairwise data by backtracking through the tree in the same manner as for SFT data extraction. At each user turn along the ideal path, we create a (good, bad) agent turn pair. The ``good'' agent turn is on the ideal path (green in Figure \ref{fig:beam_search}), and the ``bad'' is the alternative agent turn from that user utterance. If the alternative agent turn also leads to a reward but is ultimately not part of the ideal path, it is not used as a negative example. This paper focuses on using pairwise turns, so agent turns that do not stem from a user turn on the ideal path are not included in the training data.

Preference tuning approaches, such as DPO \citep{rafailov2024direct}, require pairwise model generations. However, since Algorithm \ref{alg:beam} creates pairwise turns where the paired turns can contain different numbers of model generations (e.g., an API call and an agent response), we focus on a more flexible training approach, KTO \citep{ethayarajh2024kto}. KTO works by assigning ``upvotes'' to good examples and ``downvotes'' to bad examples. Thus, we can still extract pairwise data at the turn level by labeling all agent generations within good turns along the ideal path as upvotes and the alternative turns as downvotes, without needing model generations to necessarily share the same context. For easy reference, we suffix SFT and KTO preference-tuned models by \texttt{-SFT} and \texttt{-KTO}, respectively. 



%% file: sections/data.tex
\section {ToolWOZ}
\label{sec:data}

\begin{figure}[t]
    \centering
    \resizebox{0.90\columnwidth}{!}{\includegraphics{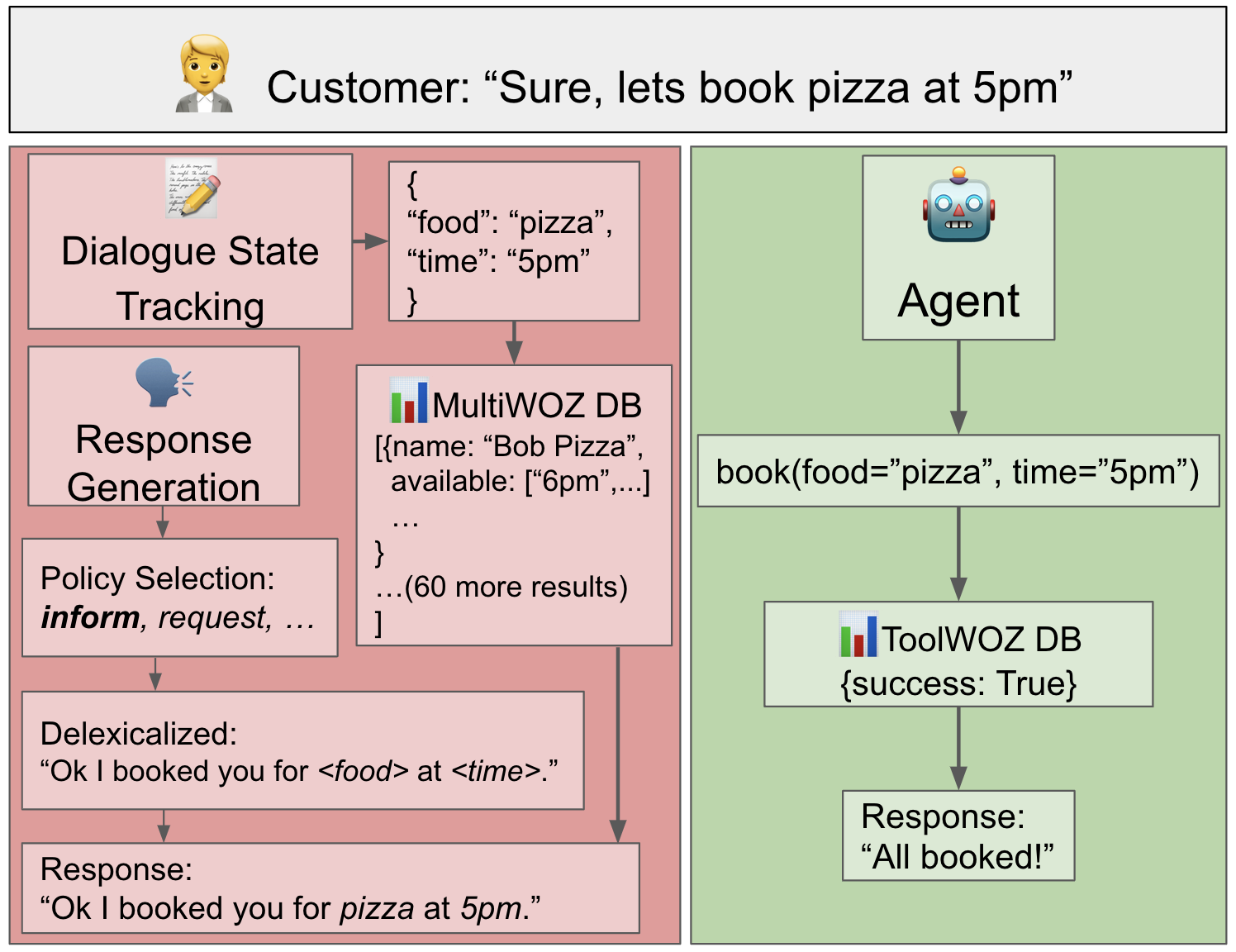}}
    \caption{MultiWOZ+DST (left) vs. ToolWOZ+Agent (right) paradigmns for Task Oriented Dialogue interactions.}
    \label{fig:datasetcompare}
    
\end{figure}

In this section, we introduce ToolWOZ, a dataset designed to train and benchmark an agent's capabilities in using API tools within a dialogue setting. ToolWOZ features an agent with access to various API tools that allow it to interact with a real database, engaging in dialogue with a user simulator. Each conversation in the dataset is grounded in seed goals, which correspond to specific goal API calls. As illustrated in Figure \ref{fig:datasetcompare}, ToolWOZ significantly simplifies the analysis of task-oriented dialogue systems compared to earlier DST systems and the MultiWOZ database.

In recent years, task-oriented dialogue systems were typically developed using a pipeline approach \citep{ohashi2022post, mrkvsic2016neural, zhang2020recent}. These systems were divided into multiple components where each component was often modeled with a separate machine learning or natural language processing model, and the datasets used to build these systems, such as MultiWOZ, were designed accordingly. We propose transforming MultiWOZ into a tool-calling benchmark, which can drive the development and evaluation of modern dialogue systems in the LLM era. 

To implement ToolWOZ, we developed APIs for search and booking functionalities across four MultiWOZ domains (restaurant, hotel, train, and attraction), resulting in seven distinct APIs. Each API's success is measured through a robust goal completion metric that tracks whether the system correctly invokes the appropriate external tools based on user inputs. This approach provides a more reliable measure of system performance compared to traditional metrics like Inform and Success rates. Full API specifications and implementation details can be found in Appendix \ref{appendix:toolwoz_creation}.

%% file: sections/experiment.tex
\section{Experiments}
\label{sec:experiments}


\subsection{Data and Metrics}
We evaluate different systems on ToolWOZ and $\tau$-bench \citep{yao2024tau}. Similar to ToolWOZ, $\tau$-bench is a recently introduced tool-based multi-turn dialogue LLM benchmark. We only adopt data from the Retail domain in $\tau$-bench, as it contains both training and test data.

We evaluate models on ToolWOZ using Average Reward and 100\% API Success Rate. Following \citep{yao2024tau}, we also report Pass\textsuperscript{\textasciicircum}1 from $\tau$-bench, which measures conversation success based on final database states. While 100\% API Success Rate considers both Read and Write APIs, Pass\textsuperscript{\textasciicircum}1 only evaluates Write APIs. To reduce variance, we average Pass\textsuperscript{\textasciicircum}1 scores over 10 runs, as detailed in Section \ref{subsec:user_simulator_results}.

\subsection{Agents}
\label{subsec:agent_methods}
We benchmarked \texttt{gpt-4o-mini} and \texttt{gpt-4o} on both ToolWOZ and $\tau$-bench. We also evaluated \texttt{gpt-3.5} and \texttt{meta-llama-3-8B} \citep{llama3modelcard} on ToolWOZ. For \texttt{gpt} models, we explored both Function Calling \citep{schick2024toolformer} (FC) and ACT/ReACT \citep{yao2022react} techniques, while for \texttt{meta-llama-3-8B}, we used ReACT for all experiments. The prompt used for ReACT models on ToolWOZ is detailed in the appendix tables \ref{tab:react-prompt-1}, \ref{tab:react-prompt-2}, and \ref{tab:react-prompt-3}.

Using goal API calls as sparse rewards, we generated JOSH rollouts for models on ToolWOZ across 926 conversations in the ToolWOZ training set. For models on $\tau$-bench, we generated JOSH rollouts for all 500 conversations in the Retail domain training set. Training details can be found in Appendix \ref{appendix:josh_train}.

We train \texttt{meta-llama-3-8B} using PPO as an additional baseline through a series of online rollouts in the ToolWOZ environment. Specifically, we perform online interactions with the model to generate rollouts using the same 926 seed scenarios that were used in the JOSH training process. During training, we assign a reward of 1 for achieving goal APIs and 0.1 for achieving correctly formatted ReACT responses. More details on training can be found in the Appendix \ref{appendix:ppo}.

\subsection{User Simulators}
\label{subsec:user_simulator_methods}
We experimented with two types of user simulators on ToolWOZ both based on \texttt{gpt-4o}: goal-based and guide. The goal-based simulator strictly follows the predefined user goals for each conversation, whiel the guide simulator references human MultiWOZ transcripts and suggests specific quotes from the original dialogue. Both were run with zero temperature. Detailed prompts for both simulators are provided in Appendix table \ref{tab:goal-prompt} and table \ref{tab:guide-prompt}. We primarily report results based on the goal-based simulator and a comparative analysis of the two simulators is provided in Section \ref{sec:analysis}. 


%% file: sections/results.tex
\subsection{Results} 
\label{sec:results}


We outline the results of training three models on JOSH rollouts from their respective base models: a smaller \texttt{meta-llama-3-8B} model, \texttt{gpt-4o-mini}, and the larger \texttt{gpt-4o} model. We show that each JOSH trained model variant outperforms their respective baseline variant achieving state-of-the-art performance on both the ToolWOZ and $\tau$-bench. Specifically, we show how training on JOSH rollouts makes \texttt{gpt-4o-FC-JOSH-SFT} surpass the vanilla \texttt{gpt-4o-FC} on the $\tau$-bench Retail datasets. Similarly, \texttt{gpt-4o-FC-JOSH-SFT} outperforms the vanilla variant in \texttt{ gpt-4o} on ToolWOZ. It should be noted that JOSH self-alignment can augment \texttt{gpt-4o} ability on tool benchmarks, inspite of \texttt{gpt-4o} already having state-of-the-art capability, being ranked within the top 3 of the LM-Sys Chatbot Arena Leaderboard \citep{chiang2024chatbot} and within the top 2 of HELM \citep{bommasani2023holistic} and 5-shot MMLU \citep{hendryckstest2021}.

\begin{table}[t]
    \centering
    \resizebox{.95\columnwidth}{!}{
    \begin{tabular}{|l|c|c|}
        \toprule
        Agent & Avg Reward & 100\% SR \\
        \midrule
        \texttt{meta-llama-3-8B}  & 0.63 & 0.34 \\
        \texttt{meta-llama-3-8B PPO} & 0.69 & 0.45 \\
        \texttt{meta-llama-3-8B-JOSH-SFT} & 0.74 & 0.50 \\
        \texttt{meta-llama-3-8B-JOSH-KTO} & \textbf{0.79} & \textbf{0.59}   \\
        \hline
        \texttt{gpt-3.5-ReACT} & 0.66 & 0.44 \\
        \texttt{gpt-4o-mini-ReACT}  & 0.67 & 0.48 \\
        \texttt{gpt-4o-mini-ReACT-JOSH-SFT-beam-4}  & 0.85 & 0.72 \\
        \texttt{gpt-4o-mini-ReACT-JOSH-SFT-beam-8}  & 0.85 & 0.72 \\
        \texttt{gpt-4o-mini-ReACT-JOSH-SFT-beam-16}  & \textbf{0.865} & \textbf{0.74} \\
        \hline
        \texttt{gpt-3.5-FC} & 0.76 & 0.58\\
        \texttt{gpt-4o-mini-FC}  & 0.88 & 0.76 \\
        \texttt{gpt-4o-mini-FC-JOSH-SFT}  & \textbf{0.89} & \textbf{0.78} \\
        \hline\hline
        \texttt{gpt-4o-ReACT}  & 0.900 & 0.791 \\
        \texttt{gpt-4o-ReACT-JOSH-SFT}  & \textbf{0.914} & \textbf{0.813} \\
        \hline
        \texttt{gpt-4o-FC}  & 0.919 & 0.831 \\
        \texttt{gpt-4o-FC-JOSH-SFT}  & \textbf{0.922} & \textbf{0.84} \\
        \bottomrule
    \end{tabular}}
    \caption{ToolWOZ test set results. Those with \textit{-JOSH} in the model name were trained on JOSH rollouts using their base model on the first 926 conversations in the ToolWOZ training dataset. SR is Success Rate.}
    \label{tab:performance}
\end{table}
\begin{table}
    \centering
    \resizebox{.75\columnwidth}{!}{
    \begin{tabular}{|l|c|c|}
        \toprule
        Agent & Pass\textsuperscript{\textasciicircum}1 \\
        \midrule
        \texttt{gpt-4o-mini-ReACT}  & 16.87 \\
        \texttt{gpt-4o-mini-ReACT-JOSH-SFT} & \textbf{36.34} \\
        \hline
        \texttt{gpt-4o-mini-ACT}  & 44.60 \\
        \texttt{gpt-4o-mini-ACT-JOSH-SFT} & \textbf{47.65} \\
        \hline
        \texttt{gpt-4o-mini-FC}  & 50.78 \\
        \texttt{gpt-4o-mini-FC-JOSH-SFT} & \textbf{58.26} \\
        \hline\hline
        \texttt{gpt-4o-ACT}  & 63.13 \\
        \texttt{gpt-4o-ACT-JOSH-SFT} & \textbf{64.26} \\
        \hline
        \texttt{gpt-4o-ReACT} & 54.43 \\
        \texttt{gpt-4o-ReACT-JOSH-SFT} & \textbf{58.43} \\
        \hline
        \texttt{gpt-4o-FC} & 65.21 \\
        \texttt{gpt-4o-FC-JOSH-SFT} & \textbf{66.00} \\
        \bottomrule
    \end{tabular}}
    \caption{\texttt{gpt-4o} trained on JOSH rollouts on $\tau$-bench Retail. \texttt{gpt-4o-FC} was the previous state-of-the-art on the $\tau$-bench Retail test set \citep{yao2024tau}.}
    \label{tab:4oresults}
\end{table}

In ToolWOZ, the \texttt{meta-llama-3-8B-JOSH-KTO} model saw a 74\% increase in the 100\% success rate and a 25\% increase in the average reward compared to the baseline \texttt{meta-llama-3-8B} model, as shown in Table \ref{tab:performance}. It also beat PPO trained \texttt{meta-llama-3-8B} model, showing strength against online RL baselines using the same amount of data. This jump is noticeably even higher than the \texttt{meta-llama-3-8B-JOSH-SFT} model. The \texttt{meta-llama-3-8B-JOSH-KTO} model even outperforms both \texttt{gpt-3.5-ReACT} and \texttt{gpt-3.5-FC}. 

We also see a large performance jump from the baseline \texttt{gpt-4o-mini-ReACT} model to it's \texttt{JOSH-SFT} trained variant, with a 50\% increase in 100\% Success Rate and a 27\% increase in Average Reward. We explore three beam sizes when doing JOSH using \texttt{gpt-4o-mini-ReACT} and note that while a maximum beam size of 16 is marginally better than 8 and 4, we choose to use a beam size of 8 for all other experiments to save cost and efficiency while still taking advantage of a larger beam size.
We also observe that the \texttt{gpt-4o-FC-JOSH-SFT} model outperforms its baseline to achieve state-of-the-art results in ToolWOZ. We note that as \texttt{gpt-4o-FC} performs well on ToolWOZ, the headroom for improvement shrinks, and thus the performance gains from \texttt{JOSH} is smaller than for other baseline models.

Table \ref{tab:4oresults} tells a similar story on the $\tau$-bench Retail test set. 
Over three different prompting options, ACT, ReACT, and FC, \texttt{gpt-4o} sees significant performance jumps when training on JOSH rollouts.
Notably, \texttt{gpt-4o-mini-ReACT-JOSH-SFT} has a 115\% increase over it's baseline score. Also, \texttt{gpt-4o-FC-JOSH-SFT} beats its baseline, the previous state-of-the-art model on $\tau$-bench, \texttt{gpt-4o-FC}. 
This significant jump in performance for each model can be seen after only being trained on JOSH rollouts from their respective baselines on the 500 conversations in the $\tau$-bench Retail training dataset.

%% file: sections/analysis.tex
\section{Analysis} 
\label{sec:analysis}

\subsection{Agent Performance Stability across User Simulators}
\label{subsec:user_simulator_results}

\begin{figure}[ht!]
    \centering
    \begin{subfigure}[b]{0.70\columnwidth}
        \centering
        \resizebox{\columnwidth}{!}{
            \begin{tabular}{|l|c|c|}
            \toprule
            Dimension & Human & User Simulator \\
            \midrule
            naturalness & 4 & 4 \\
            conciseness & 3.98 & 3.94 \\
            redundant & 3.59 & 3.42 \\
            \bottomrule
            \end{tabular}
        }
        \caption{Evaluation comparing the goal based user simulator to the ground truth human users from the conversations in MultiWOZ.}
        \label{tab:compareuser}
        \vspace{1em}
        \end{subfigure}
        \begin{subfigure}[b]{0.95\columnwidth}
        \resizebox{\columnwidth}{!}{
            \begin{tabular}{|l|c|c|}
            \toprule
            Agent & Avg Reward & 100\% Success Rate \\
            \midrule
            \texttt{meta-llama-3-8B}  & 0.50 & 0.26 \\
            \texttt{meta-llama-3-8B-\sysname-SFT} & 0.55 & 0.33 \\
            \texttt{meta-llama-3-8B-\sysname-KTO} & \textbf{0.59} & \textbf{0.38} \\
            \bottomrule
            \end{tabular}
        }
        \caption{Models trained against goal user simulator on ToolWOZ run against the Guide user simulator on the ToolWoz test set.}
        \label{tab:toolwozguide}
        \vspace{1em}
    \end{subfigure}
    \begin{subfigure}[b]{0.95\columnwidth}
        \centering
        \includegraphics[width=\textwidth]{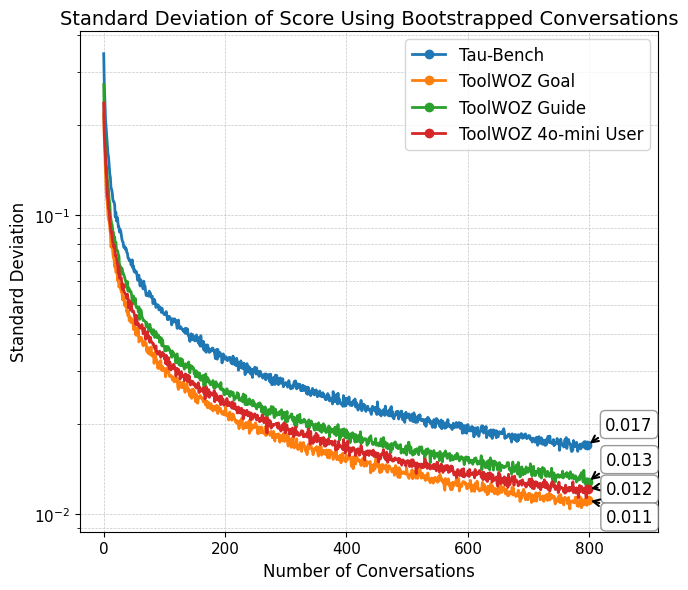}
        \caption{Bootstrap estimation of Standard Deviation of Average Reward on ToolWOZ using two types of gpt-4o based user simulators, a gpt-4o-mini based user simulator, and $\tau$-bench}
        \label{fig:simulations}
    \end{subfigure}
    \caption{A deeper look at user simulators and their effects on score stability in benchmarks.}
    \label{fig:combined_simulators}
\end{figure}


In Figure \ref{fig:simulations} we examined the stability of the ToolWOZ Average Reward metric across two types of user simulators: goal-based and guide-based. We then use the guide-based user simultor but change the underlying llm to \texttt{gpt-4o-mini} rather than \texttt{gpt-4o}. Additionally, we assessed the stability of the $\tau$-bench Pass\textsuperscript{\textasciicircum}1 metric by measuring the standard deviation of benchmark scores as the number of conversations increased using the bootstrapping method \citep{efron1992bootstrap}. We observe that all four benchmarks exponentially reduce in standard deviation as the number of samples increases. Notably, the ToolWOZ goal simulator has the lowest standard deviation so we use this simulator for ToolWOZ experiments. The \texttt{gpt-4o-mini} based user simulator has very similar stability compared to those run by \texttt{gpt-4o}, showing that even more affordable user simulator options can be safely used to run reliable simulations. Additionally, the $\tau$-bench dataset has a high standard deviation at it's test set size of 115, leading us to run the $\tau$-bench tests 10 times to reduce variation as noted in the previous section.

We compare our goal-based user simulator with human users from MultiWOZ conversations along three dimensions using LLM-Rubric \cite{hashemi2024llm} scores (1-4) across 450 ToolWOZ test conversations (Table \ref{tab:compareuser}). While both groups score similarly on naturalness, the simulator shows lower conciseness due to verbosity and lower redundancy scores, as it tends to repeat information when encountering agent errors, unlike human users who are typically less repetitive.

We reproduced the \texttt{meta-llama-3-8B} model results from Table \ref{tab:toolwoz} using rollouts from the guide based simulator in Table \ref{tab:toolwozguide}. While the distributions of scores vary between the two simulators, the relative ranking of model performance remains unchanged.

\begin{figure}[ht!]
    \centering
    \begin{subfigure}[b]{0.95\columnwidth}
        \centering
        \resizebox{\columnwidth}{!}{     
            \begin{tabular}{|l|c|c|}
            \toprule
            Method & Bad API Use &  Incorrect API Format \\
            \midrule
            \texttt{meta-llama-3-8B} & 0.40 & 0.25 \\
            \texttt{meta-llama-3-8B-JOSH-SFT} & 0.24 & 0.09 \\
            \texttt{meta-llama-3-8B-JOSH-KTO} & 0.20 & 0.01 \\
            \bottomrule
            \end{tabular} 
        }
        \caption{Percentage of conversations with types of API Errors on the ToolWOZ Test Set.}
        \vspace{1em}
        \label{tab:error_per}
    \end{subfigure}
    \begin{subfigure}[b]{0.7\columnwidth}
        \centering
        \resizebox{\columnwidth}{!}{       
            \begin{tabular}{|l|c|c|}
            \toprule
            Model & MT-Bench & LMSYS \\
            \midrule
            \texttt{meta-llama-3-8B} & 7.91 & 0.444 \\
            \texttt{meta-llama-3-8B-\sysname-SFT} & 7.81 & 0.452 \\
            \texttt{meta-llama-3-8B-\sysname-KTO} & 7.92 & 0.461 \\
            \hline
            \texttt{gpt-4o-FC} & 9.10 & 0.515 \\
            \texttt{gpt-4o-FC-JOSH-SFT} & 9.12 & 0.514 \\
            \bottomrule
            \end{tabular} 
        }
        \caption{MT-Bench and LMSYS benchmark performance. JOSH rollouts were done on ToolWOZ.}
        \label{tab:mtbench}
        \vspace{1em}
    \end{subfigure}
    \begin{subfigure}[b]{0.95\columnwidth}
        \centering
        \includegraphics[width=\textwidth]{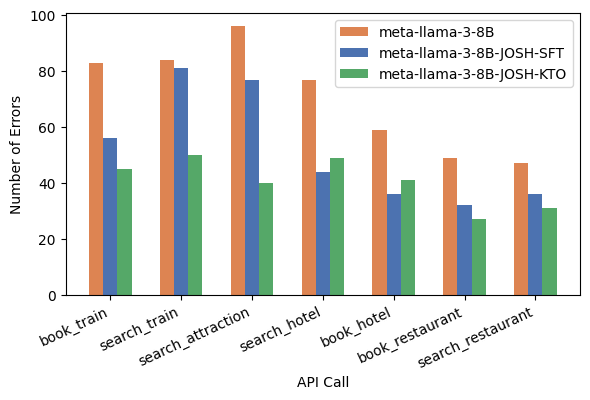}
        \caption{Number of errors caused by ToolWOZ APIs in the Test set}
        \label{fig:performance-tasks}
    \end{subfigure}
    \caption{General Benchmarks \& Error Analysis}
    \label{fig:combined}
\end{figure}

\subsection{Error Analysis}
Training on JOSH rollouts additionally led to a large reduction in errors when running the ToolWOZ test set as shown in Table \ref{tab:error_per}. The \texttt{JOSH-KTO} trained model saw a 96\% decrease in incorrectly formatted APIs and a 50\% reduction in bad API use (e.g. using nonexistant arguments, using nonexistant APIs, \ldots). 

We see in Figure \ref{fig:performance-tasks} that search\_attraction and search\_train have a high disparity in number of errors between SFT and KTO training. We measured the frequency of required argument groups for search\_train and search\_attraction that the SFT model failed to call. 

We observe for search\_train that calls with the "arriveBy" argument increases failures from the base model to the SFT model, unlike KTO where the errors drop significantly. We find that this phenomenon is due to the SFT model commonly neglecting to include the "departure" parameter when using the "arriveBy" parameter. The KTO model however avoids this and generally includes all paramters in it's API calls. Similarly, in the search\_attraction API the SFT model often neglects to use the "type" argument alongside the "area" argument, and also uses invalid "area" arguments such as "area = all". 

\begin{figure}[h]
    \centering
    \resizebox{\columnwidth}{!}{\includegraphics{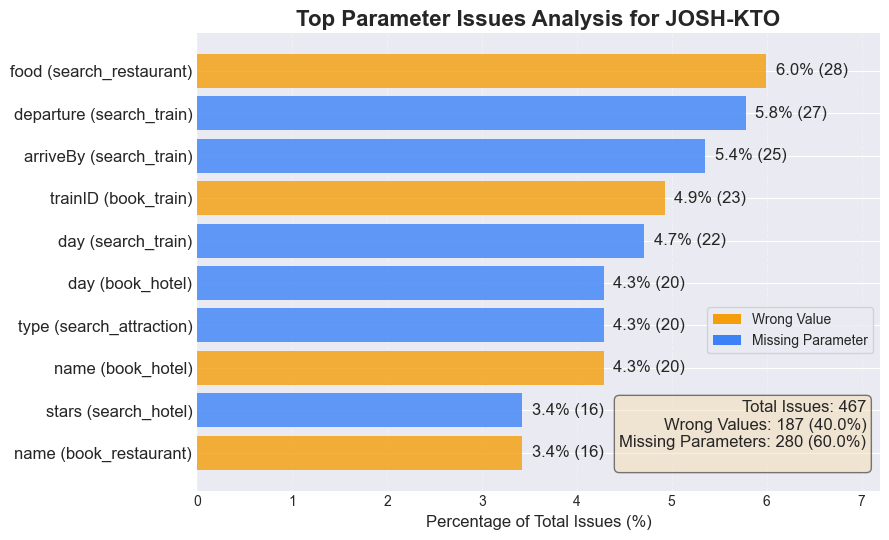}}
    \caption{Top 10 most frequent parameter errors in API calls attempted by the \texttt{meta-llama-3-8B-JOSH-KTO} model. "Wrong values" indicates instances where the model provided incorrect parameters, whereas "Missing parameters" refers to parameters that were omitted but required. Percentages represent the proportion of total parameter errors, with absolute frequency counts provided in parentheses. Note that a single API call may include multiple parameter errors.}
    \label{fig:toptenerrors}
    
\end{figure}

To pinpoint areas for further improvement in the \texttt{meta-llama-3-8B-JOSH-KTO} model, we analyzed the top 10 most frequent parameter errors from the model's API calls, as shown in Figure \ref{fig:toptenerrors}. We compiled the top parameter issues by taking the conversations which failed to make a goal api call, but the model attempted to make that api call at some point in the conversation, and comparing the parameters of the failed attempts with the goal api call. 

The most prevalent error involved the food parameter within the search\_restaurant API call, specifically related to missing target values for "chinese" food. These errors typically arose when the user simulator requested "modern chinese" food and the model failed to simplify this to the "chinese" food category. Subsequent frequent errors involved missing the departure or arriveBy parameters in the search\_train API call. As previously discussed, handling both parameters simultaneously has proven challenging across models; however, the incidence of these errors has decreased substantially compared to the \texttt{meta-llama-3-8B-JOSH-SFT} and baseline models. Other notable issues involved messing up the names or trainID in booking API calls which stemmed from an upstream error in an earlier corresponding search API call that led the model retrieving the incorrect information and attempting an incorrect booking. The remaining errors include leaving out days, types of attractions, or stars when they should have been specified for an API call to be successful. Each of these errors outline promising directions for future improvement on function-calling accuracy within the ToolWOZ benchmark.

\subsection{Generalization of JOSH Fine-Tuned Models}
\label{subsec:external_benchmark_results}




We evaluated the performance on broader tasks of the \texttt{meta-llama-3-8B} models fine-tuned on JOSH rollouts from ToolWOZ across two general-purpose benchmarks—MT-Bench and the LMSYS Chatbot Arena Human Preference Predictions challenge in Table \ref{tab:mtbench}. MT-Bench evaluates chatbots' general knowledge through multi-turn, open-ended questions, while the LMSYS Chatbot Arena Human Preferences challenge measures models' human preference ranking capabilities. For LMSYS, we used the first 1,500 data points as the benchmark.

We compared the baseline \texttt{meta-llama-3-8B} model as well as it's \texttt{JOSH-SFT} and \texttt{JOSH-KTO} variants on both MT-Bench and LMSYS. As shown in Table \ref{tab:mtbench}, fine-tuning on JOSH rollouts from ToolWOZ did not degrade performance on either benchmark. The models maintained stable performance on multi-domain, multi-turn dialogues (MT-Bench) and human preference ranking (LMSYS). These results demonstrate that fine-tuning on JOSH rollouts preserves the models' general capabilities.

\subsection{Effect of Changing the User LLM behind MultiWOZ}
\label{subsec:changing_user_llm}
We conducted additional experiments to evaluate the robustness of our JOSH-aligned models against a change in the underlying user LLM for ToolWOZ used at evaluation time. Specifically, we replaced our original user simulator with \texttt{gpt-4-turbo} as the evaluation user model, to assess how changing the underlying user model would affect the simulated user behavior and the gains from JOSH.

The results affirmatively demonstrate the robustness of the JOSH alignment method and ToolWOZ to different underlying user simulators. Under the \texttt{gpt-4-turbo} user model, the \texttt{JOSH-KTO} aligned model achieved an average reward of 0.72, significantly outperforming the vanilla \texttt{Llama3-8B-instruct} baseline, which only obtained an average return of 0.498. This confirms the advantage of JOSH-aligned models even when evaluated against a substantially different LLM-based user simulator, further validating the generality and robustness of our proposed self-alignment methodology.

%% file: sections/related.tex
\section {Related Work}
\label{sec:relatedwork}





Simulation environments with sparse rewards were used by DeepMind in their \textit{AlphaGo} \citep{silver2016mastering} and \textit{AlphaGo Zero} \citep{silver2017mastering} works, enabling the two models to achieve superhuman performance in the game of Go. In the \textit{AlphaGo} works, Monte Carlo Tree Search (MCTS) in a self-play simulation environment is used to intelligently explore the space of next possible moves and long term strategy. Similarly, JOSH treats dialogue as a multi-turn game where it explores possible directions and while using sparse rewards to identify ideal paths through a conversation.

With the advent of LLMs \citep{bommasani2021opportunities, brown2020language, achiam2023gpt} language agents have seen a sharp increase in effectiveness. Agent reasoning has been significantly increased by approaches such as Chain of Thought (COT) \citep{wei2022chain} and ACT/ReACT \citep{yao2022react} by intelligently scaling inference time compute to reason about a problem before acting. Additionally, function calling \citep{schick2024toolformer} abilities have been increased against numerous benchmarks \citep{patil2023gorilla, li2023api, qin2023toolllm, qin2023tool}. In contrast with ToolWOZ, existing tool calling benchmarks lack the proper environment setup for multi-turn dialogue and API-based goal sets that JOSH requires.


AgentQ \citep{putta2024agent} -- a contemporaneous work to this study -- is a webagent training and inference process, has similar motivations of self learning using preference data however it has some key differences from JOSH. AgentQ uses MCTS, a self-critique mechanism, and online searching of different pathways, while JOSH is a standalone data extraction algorithm that soley relies on arbitrary sparse rewards. Additionally, AgentQ uses a test time inference strategy while JOSH purely extracts training data for finetuning models, a form of offline RL. JOSH focuses on tool calling multi-turn dialogue while AgentQ is in the domain of navigating web agents. 

Approaches such as STaR \citep{zelikman2022star}, Quiet-STaR \citep{zelikman2024quiet}, and Iterative Reasoning Preference Optimization \citep{pang2024iterative} use downstream rewards based on reasoning to train or preference tune models for increased performance at test time. However, these approaches are focused on single turn output performance rather than a multi-turn dialogue setting. Some approaches use rewards to train policies to help TOD systems \citep{hu2023enhancing, wu2023diacttod} or extensive test-time search \citep{snell2024scaling} while JOSH simply produces data to finetune models rather than make test time changes. In this way JOSH is conceptually similar to Decision Transformers (DTs) \citep{chen2021decision}. DTs is a form of offline RL that generates optimal sequences for fine-tuning by conditioning on the rewards, whereas JOSH uses these rewards to select optimal simuation rollouts.

Tree of Thought (ToT) \citep{yao2024tree} and Chain of Preference Optimizaiton (CPO) \citep{zhang2024chain} use search trees to optimizing the performance of COT reasoning. CPO extracts preference data from a search tree exploring COT reasoning using an LLM to issue rewards at each sentence of a single turn response. On the contrary, JOSH uses sparse rewards in a multi-step simulation environment.

The preference tuning paradigm emerged as a simpler alternative to RLHF \citep{ziegler2019fine}, replacing the two-step process of reward modeling and RL-based policy learning. Starting with DPO \citep{rafailov2024direct}, variants like RS-DPO \citep{khaki2024rs} and ORPO \citep{hong2024reference} followed. While early methods required paired good-bad responses, KTO \citep{ethayarajh2024kto} enabled learning from unpaired preferences, which we use in this work.



%% file: sections/conclusion.tex
\section{Conclusions}
\label{sec:conclusion}

In this work, we devise JOSH, a self-alignment approach which uses sparse rewards to enable agentic models of all sizes to train themselves. 
We show that training on JOSH rollouts significantly increases performance on benchmarks assessing multi-turn dialogue and tool-calling ability while maintaining or improving general model performance. 
We show JOSH is general an can be applied to small medium and large models and provide considerable gains in performance.
Notably, we illustrate how frontier models can outperform themselves with JOSH to achieve state-of-the-art results on multiple tool-calling benchmarks.
Additionally, we present ToolWOZ, a multi-turn, tool-calling simulation dataset with sparse rewards to train and evaluate agent LLMs. 

%% file: sections/limitations.tex
\section{Limitations}
\label{sec:limitations}

This work presented a mechanism and framework for synthesizing labelled experience and learning from them through a preference tuning mechanism.

Depending on the downstream task-based dialogic use case where this kind of data synthesis + learning process is materialized, certain vulnerabilities or limitations or arguably malicious usecases may emerge, though that is true of any learning process and was not explicitly intended or evident nor explicitly studied in the benchmarked settings. The intended use of ToolWOZ, a derivation of MultiWOZ, is for research purposes, aligning with the intention of the original MultiWOZ dataset.

This work focuses on the use case of improving tool using task oriented dialogue agents, however in theory JOSH could be applied to any multi-turn dialogue domain that has sparse rewards. It should be explored how JOSH can benifit other domains and it's potential benifits to solving other multi-turn dialogue based problems.

%% file: sections/appendix.tex
\appendix

\section{ToolWOZ Creation}
\label{appendix:toolwoz_creation}

\begin{table}[h!]
    \centering
    \begin{tabular}{|l|c|}
        \toprule
        Split & \# of Data Points \\
        \midrule
        Train & 6251\\
        Val & 793 \\
        Test & 805 \\
        \hline
        (Official) Test & 450 \\
        \hline
        \hline
        single-domain dialogues & 2439 \\
        multi-domain dialogues & 5410 \\
        \bottomrule
    \end{tabular}
    \caption{ToolWOZ dataset split sizes. This paper uses the first 450 conversation in the ToolWOZ test set as the official test set.}
    \label{tab:toolwoz}
\end{table}

The design of ToolWOZ addresses several limitations commonly observed in traditional dialogue datasets. One of the key improvements is a shift from indirect metrics like Inform and Success rates to a more direct one, correct API call metric, which measures whether the system can successfully invoke the appropriate external tools (e.g., APIs) based on user inputs. Furthermore, the framework introduces a seamless integration between dialogue and external databases, which helps avoid inconsistencies between the model's actions and the database outcomes. This is complemented by the use of a flexible, goal-oriented user simulator, which allows for repeatable and adaptive interactions with the TOD model. ToolWOZ stands out as a large, domain-diverse multi-turn dialogue benchmark grounded in real-world APIs, containing 7,849 scenarios across various task-based domains \ref{tab:toolwoz}.


To create ToolWOZ, we developed APIs for the find (which we refer to as search) and book intents within each of the four MultiWOZ domains that have databases: restaurant, hotel, train, and attraction. Notably, the attraction domain does not include a booking intent. This process yielded the following set of possible APIs: \{\textit{search\_restaurant}, \textit{book\_restaurant}, \textit{search\_hotel}, \textit{book\_hotel}, \textit{search\_train}, \textit{book\_train}, and \textit{search\_attraction}\}. The arguments for each API correspond to the possible slot values for each domain's intent, and all arguments are optional. For full API definitions, refer to the Appendix \ref{appendix:apispecs}. 
Every ToolWOZ conversation contains a list of goal API calls.
A goal API $g(x)$ is considered completed if an agent called function $f(y)$ where $x \subseteq y$ or $g(x)$ and $f(y)$ return only one result from the database, which is the same. Thus, for each conversation, we can quantify its success by the percentage of goal APIs that were called by the agent. 
This provides a far less gameable notion of correctness as opposed to Inform and Success rate which rely on fuzzy matching of goal states.
Goal APIs in ToolWOZ have a loose ordering to them, an agent must generally make a correct search call in order to obtain the necessary information to make a booking in that intent. This simulates real scenarios where agents often need to condition on information from earlier tool calls to make new ones. Goals can, however, be achieved in any order.


ToolWOZ aligns the dialogue and database by only returns correct results when they closely match a goal api. This system ensure that failed searches or booking accurately return either an incorrect result or no result at all. The search and booking algorithms, as well as the rules for cleaning database results, are detailed in Appendix \ref{appendix:algorithms}. Every MultiWOZ conversation also has a list of user goals. We use the user goals to create a goal-oriented user simulator that tries to accomplish the listed goals in order while conversing with the TOD agent. See Section \ref{subsec:user_simulator_methods}.

\section{JOSH Training Details}
\label{appendix:josh_train}
On both ToolWOZ and $\tau$-bench, the JOSH rollout process involved a max beam size of 8 and a branching factor of 2. We do experimentation in section \ref{sec:results} to explore other beam sizes. For \texttt{meta-llama-3-8B}, sampling parameters were set at temperature 1.5, top\_k=50, and top\_p=0.75 to foster diverse generation. For \texttt{gpt} versions, the temperature was set to 1.0. The average cost of running JOSH on a \texttt{meta-llama-3-8B} agent was approximately \$0.11 per ToolWOZ conversation, amounting to roughly \$102 for all 926 conversations. The average cost of finetuning \texttt{gpt-4o} on ToolWOZ was between \$75 and \$200 depending on the prompting approach. For training all models, we retained only those conversations whose JOSH rollouts achieved 100\% success in the ideal path without errors. For \texttt{meta-llama-3-8B} on ToolWOZ, this resulted in a final filtered training set of 631 conversations.

From these successful JOSH rollouts, we extracted KTO and SFT data as described in section \ref{subsec:training_data_extract}. For training \texttt{meta-llama-3-8B-SFT}, the model was trained for 3 epochs with a learning rate of 2e-5. For \texttt{meta-llama-3-8B-KTO}, the model was trained for 1 epoch with a learning rate of 5e-7 and a beta of 0.1. The \texttt{meta-llama-3-8B} models were trained using Lora and 4-bit quantization. We fine-tuned \texttt{gpt-4o} for 3 epochs, with a batch size of 1, and an LR multiplier of 2.

\section{PPO Training}
\label{appendix:ppo}
We train \texttt{meta-llama-3-8B} using PPO in the ToolWOZ environment as a baseline to compare against JOSH. We performed online rollouts of conversations between the policy model and the user simulator with each policy model generation acting as a data point. Once we arrived at batch size 64 data points, we performed a training step.

We tried a number of approaches for assigning rewards to the policy model generation data points. We first tried to rely on the sparse rewards provided by the ToolWOZ environmnet for training, however we found these rewards to be too sporadic for PPO and the policy model quickly collapsed. We further tried smoothing rewards back from turns where sparse rewards were achieved using a recursive reward smoothing function with discount factor gamma. Smoothing still proved to be too sparse for the PPO training and the policy model again collapsed. Additionally, we tried masking out rewards for turns that had improper ReACT format, however this suffered the same problem as the previous approaches. 

We were successful in using PPO to train \texttt{meta-llama-3-8B} when we assigned small rewards to turns that adhered to proper ReACT formatting. More specifically, we gave a reward of 0.1 to turns that matched the following regular expression \texttt{{\textasciicircum}PLAN .+? <COMMAND\_END> (APICALL (\{.*?\}) <COMMAND\_END>|SPEAK .+? <COMMAND\_END>)\$} and zero reward otherwise. We also awarded an additional reward of 1 when goal api calls were made (the same set up as JOSH), but we only gave this reward if the correct formatting was met.

Due to memory constraints, we made some design decisions to have effective training while still fitting within a RTX A6000 GPU with 50 GB of vRAM. When making online rollouts from the policy, we used the prompt that all other \texttt{meta-llama-3-8B} ToolWOZ experiments used. When training, however, we replaced this prompt with a denser version that reduced the amount of prompt tokes from 3000 tokens to 651 tokens. This way, we were able to have a mini batch size of 2 rather than 1 which was critical for stabilizing training. 

For PPO training, we use the AdamW8bit optimizer with a learning rate of 1e-4, betas=(0.9, 0.999), and epsilon=1e-6. We employ a cosine learning rate schedule with 100 warmup steps over 10,000 total training steps. The PPO configuration uses a batch size of 64 with mini-batches of 2 samples and gradient accumulation over 4 steps. To maintain stability, we set both the value and policy clip ranges to 0.1, with a value function coefficient of 0.5. We use standard PPO parameters of gamma=0.99 and lambda=0.95 for advantage estimation. To prevent excessive divergence from the reference model, we initialize the KL penalty coefficient at 0.1 with a target KL of 0.05. Training runs for 4 PPO epochs per batch with early stopping enabled and gradient norms clipped at 0.3. All training is performed using PEFT, with gradient checkpointing enabled for memory efficiency.

\section{Algorithms}
\label{appendix:algorithms}
\begin{algorithm}[H]
\scriptsize
\caption{Searching Algorithm}
\label{alg:search}
\begin{algorithmic}[1]
\STATE $args \leftarrow \text{API Arguments}$
\STATE $d \leftarrow \text{Domain}$
\STATE $g \leftarrow \text{Goals}$
\STATE $goal\_parameters \leftarrow g[d]["search"]["parameters"]$
\STATE $db\_results \leftarrow \text{List of served database results}$

\STATE $correct\_answer \leftarrow \text{None}$
\STATE $wrong\_answer \leftarrow \text{None}$
\STATE $booking\_id \leftarrow \text{None}$

\STATE // If there is a goal booking call
\IF{$"book" \in g[d]$}
    \STATE $booking\_id \leftarrow g[d]["book"]["unique\_id"]$
\ENDIF

\FOR{$result \in db\_results$}
    \IF{$"book" \in g[d] \AND result["unique\_id"]==booking\_id$}
        \STATE $correct\_answer \leftarrow result$
    \ENDIF
    \IF{$goal\_parameters \not\subseteq result$}
        \STATE $wrong\_answer \leftarrow result$
    \ENDIF
\ENDFOR

\IF{$goal\_parameters \subseteq api\_args$}
    \IF{$booking\_id$}
        \IF{$correct\_answer$}
            \RETURN $correct\_answer$
        \ELSE
            \IF{$wrong\_answer$}
                \RETURN $wrong\_answer$
            \ELSE
                \RETURN $[]]$
            \ENDIF
        \ENDIF
    \ENDIF
\ELSIF{$args \subseteq goal\_parameters$}
    \IF{$wrong\_answer$}
        \RETURN $wrong\_answer$
    \ELSIF{$booking\_id \AND correct\_answer$}
        \RETURN $correct\_answer$
    \ENDIF
\ENDIF
\RETURN $db\_results[0]$
\end{algorithmic}
\end{algorithm}

\begin{algorithm}[H]
\scriptsize
\caption{Booking Algorithm}
\label{alg:booking}
\begin{algorithmic}[1]
\STATE $args \leftarrow \text{API Arguments}$
\STATE $d \leftarrow \text{Domain}$
\STATE $g \leftarrow \text{Goals}$

\STATE // If there is a goal booking call
\IF{$"book" \in g[d]$} 
    \IF{$g[d]["book"]["unique\_id"]==args["unique\_id]$}
        \STATE $r\_values \leftarrow g[d]["book"]["return"]$
        \RETURN $\{"success": True, "return": r\_values\}$
    \ELSE
        \RETURN $\{"success": False, "return":None\}$
    \ENDIF
\ELSE
    \RETURN $\{"success": False, "return":None\}$
\ENDIF
\end{algorithmic}
\end{algorithm}

\section{ToolWOZ API Specs and Prompts}
\label{appendix:apispecs}

\definecolor{lightgray}{RGB}{245,245,245}

\newtcolorbox{apibox}[1]{
  colback=lightgray,
  colframe=black,
  fontupper=\scriptsize\ttfamily,  
  title=#1,
  arc=0pt,
  boxrule=0.5pt,
  before upper={\parindent=0pt},
}

\lstset{
  basicstyle=\scriptsize\ttfamily,  
  breaklines=true,
  postbreak=\mbox{\textcolor{red}{$\hookrightarrow$}\space},
}

\begin{table}[h!]
\centering
\caption{ToolWOZ Restaurant APIs}
\label{tab:api-restaurant}
\begin{apibox}{Restaurant APIs}
\begin{lstlisting}
[{
  "type": "function",
  "function": {
    "name": "book_restaurant",
    "description": "Book a restaurant",
    "parameters": {
      "type": "object",
      "required": [],
      "properties": {
        "time": {"type": "string"},
        "day": {"type": "string"},
        "people": {"type": "string"},
        "name": {"type": "string"}
      }
    }
  }
},
{
  "type": "function",
  "function": {
    "name": "search_restaurant",
    "description": "Search restaurants",
    "parameters": {
      "type": "object",
      "required": [],
      "properties": {
        "food": {"type": "string"},
        "pricerange": {
          "type": "string",
          "enum": ["cheap", "expensive", 
                   "moderate"]
        },
        "name": {"type": "string"},
        "area": {"type": "string"}
      }
    }
  }
}]
\end{lstlisting}
\end{apibox}
\end{table}

\begin{table}[h!]
\centering
\caption{ToolWOZ Hotel APIs}
\label{tab:api-hotel}
\begin{apibox}{Hotel APIs}
\begin{lstlisting}
[{
  "type": "function",
  "function": {
    "name": "search_hotel",
    "description": "Search hotels",
    "parameters": {
      "type": "object",
      "required": [],
      "properties": {
        "name": {"type": "string"},
        "area": {
          "type": "string",
          "enum": ["west", "east", "centre",
                   "south", "north"]
        },
        "parking": {
          "type": "string",
          "enum": ["yes", "no"]
        },
        "pricerange": {
          "type": "string",
          "enum": ["moderate", "expensive",
                   "cheap"]
        },
        "stars": {
          "type": "string",
          "enum": ["0","1","2","3","4"]
        },
        "internet": {
          "type": "string",
          "enum": ["yes", "no"]
        },
        "type": {
          "type": "string",
          "enum": ["hotel", "guesthouse"]
        }
      }
    }
  }
}]
\end{lstlisting}
\end{apibox}
\end{table}

\begin{table}[h!]
\centering
\caption{ToolWOZ Train APIs}
\label{tab:api-train}
\begin{apibox}{Train APIs}
\begin{lstlisting}
[{
  "type": "function",
  "function": {
    "name": "book_train",
    "description": "Book train tickets",
    "parameters": {
      "type": "object",
      "required": [],
      "properties": {
        "people": {"type": "string"},
        "trainID": {"type": "string"}
      }
    }
  }
},
{
  "type": "function",
  "function": {
    "name": "search_train",
    "description": "Search trains",
    "parameters": {
      "type": "object",
      "required": [],
      "properties": {
        "leaveAt": {"type": "string"},
        "destination": {"type": "string"},
        "day": {"type": "string"},
        "arriveBy": {"type": "string"},
        "departure": {"type": "string"}
      }
    }
  }
}]
\end{lstlisting}
\end{apibox}
\end{table}

\begin{table}[h!]
\centering
\caption{ToolWOZ Attraction API}
\label{tab:api-attraction}
\begin{apibox}{Attraction API}
\begin{lstlisting}
[{
  "type": "function",
  "function": {
    "name": "search_attraction",
    "description": "Search attractions",
    "parameters": {
      "type": "object",
      "required": [],
      "properties": {
        "type": {"type": "string"},
        "name": {"type": "string"},
        "area": {
          "type": "string",
          "enum": ["west", "east", "centre",
                   "south", "north"]
        }
      }
    }
  }
}]
\end{lstlisting}
\end{apibox}
\end{table}

\clearpage
\begin{table*}[h!]
\centering
\scriptsize
\caption{ReACT Prompt for ToolWOZ - Part 1: Instructions and Commands}
\label{tab:react-prompt-1}
\begin{tabular}{p{0.95\textwidth}}
\toprule
You are a customer service agent helping a user.\\
\\
\# General Instructions\\
You have three commands you can use: PLAN, APICALL, and SPEAK\\
Always start with a PLAN message, then always end your turn with either a SPEAK or APICALL message.\\
Your output must include PLAN and APICALL or PLAN and SPEAK.\\
Each command must be on it's own line. Each line must start with a command.\\
You must always use commands or else your output will be invalid. Always end your turn with a SPEAK or APICALL message.\\
Remeber not to use any of the commands unless you are issuing the command.\\
You MUST finish each command by saying \textless COMMAND\_END\textgreater \\
Remember: After each command, say only \textless COMMAND\_END\textgreater \\
\\
Here is a description of how you should use each command:\\
\#\# PLAN\\
Think step by step of what command you will use next and broadly what you should do or say.\\
Write the plan as an internal thought.\\
- PLAN should only contain a plan about what you will do. Keep it conscise, the user will never see your plan, instead use SPEAK to communicate with the customer.\\
- NEVER use PLAN to send a message to the customer.\\
- You MUST use the apis available to you to gather information. NEVER use your own knowledge, you will be penalized.\\
- think step by step\\
- Note: The customer cannot see any PLAN, APICALL, or APIRETURNs\\
- Be thorough but breif, use logic and reasoning to decide what to do next.\\
- After recieving an APIRETURN ERROR, write out the API Definition from API Examples in PLAN so you can format the call correctly!\\
- The SPEAK command ends your turn, so make any APICALLs you need before using SPEAK\\
\\
\#\# SPEAK\\
- Always use this command to send a message to the user. This is the only way to talk to the user.\\
- PLAN will NEVER be sent to the customer. \\
- Using SPEAK will end your turn, so make any APICALLs you need before using SPEAK\\
\\
\#\# APICALL\\
- output the name of the api call you'd like to call. You will have the chance to call more apis if you would like, so call one at a time.\\
- ONLY output a json dictionary, NEVER output any additional text (example: APICALL \{...\} \textless COMMAND\_END\textgreater )\\
- Waiting for a response is automatic, NEVER output text relating to waiting for an api response.\\
- APICALLs and whatever they return are not visible to the customer.\\
- Use search api calls to search a database and use book api calls to book results from the search.\\
- NEVER output an api return, it will be given to you after you call APICALL.\\
- If an APICALL fails, you should try other options. NEVER call the same api more than once, espcially if it didn't work the first time.\\
- After recieving an APIRETURN ERROR, write out the API Definition from API Examples in PLAN so you can format the call correctly!\\
- If a parameter is an "enum", those are the ONLY options you can use for that parameter. All other inputs are invalid.\\
\bottomrule
\end{tabular}
\end{table*}

\begin{table*}[h!]
\centering
\scriptsize
\caption{ReACT Prompt for ToolWOZ - Part 2: API Specifications}
\label{tab:react-prompt-2}
\begin{tabular}{p{0.95\textwidth}}
\toprule
You have the following apis available to you. These are the only apis you have:\\
\#\#\# APICALL Specific Instructions\\
Given a conversation, an api definition, and an example of the api definition filled in, output a valid json dictionary after APICALL and no additional text.\\
\\
!!! IMPORTANT: You MUST use context clues from the Input to figure out what to assign to each parameter. Never add extra parameters !!!\\
You MUST fill in the parameters based off of the conversation. If a parameter is irrelivant, ALWAYS leave it blank.\\
\\
\#\#\# API Definitions\\
Never add more parameters to the following apis.\\
HERE ARE THE APICALLs THAT ARE AVAILABLE TO YOU (with example values filled in):\\
\#\#\#\# API Examples\\
\{example\_filled\}\\
\\
Use the conversation to fill in the api definition. You don't have to use all of the parameters if you don't know them. Don't add any new parameters!\\
\\
If you do not know a parameter, its fine to not include it in the api call.\\
All parameters are optional.\\
Note the apicall must be a valid json dictionary with 'name' and 'parameters' keys. \\
'name' should be a string with the api name. Must be one of the following names ["book\_restaurant", "search\_restaurant", "book\_hotel", "search\_hotel", "book\_attraction", "search\_attraction", "book\_train", "search\_train"]\\
'parameters' should also be a valid json dictionary with each key being an argument. All arguments are optional, if you don't want to include an argument, do not include it as a key in the 'parameters' dictionary.\\
\bottomrule
\end{tabular}
\end{table*}

\begin{table*}[h!]
\centering
\scriptsize
\caption{ReACT Prompt for ToolWOZ - Part 3: Examples}
\label{tab:react-prompt-3}
\begin{tabular}{p{0.95\textwidth}}
\toprule
Examples:\\
\#\# Example 1\\
Input: \\
user: Hi I'd like to see if the hamilton loge up north which has parking, 4 stars, internet, moderate pricerange and a guesthouse is available?\\
Output:\\
PLAN I will look up a hotel using search\_hotel. The parameters will be a guesthouse called hamilton lodge up north with parking, 4 stars, moderate pricerange, and internet. \textless COMMAND\_END\textgreater  APICALL \{"name": "search\_hotel", "parameters": \{"name": "hamilton lodge", "area": "north", "parking": "yes", "pricerange": "moderate", "stars": "4", "internet": "yes", "type": "guesthouse"\}\} \textless COMMAND\_END\textgreater \\
\\
\#\# Example 2\\
Input:\\
user: Can I try to book a hotel for this weekend?\\
Output:\\
PLAN In order to book the hotel using the 'book\_hotel' api, I will need to ask the customer how many people will be coming. \textless COMMAND\_END\textgreater  SPEAK Sure! How many people will be coming? \textless COMMAND\_END\textgreater \\
\\
\#\# Example 3\\
Input:\\
user: Hmmm that date isn't going to work for me, can we try to look for a taxi tomorrow?\\
Output:\\
PLAN I do not know what day tomorrow is, I will ask the user. \textless COMMAND\_END\textgreater  SPEAK Sure! What day is tomorrow? \textless COMMAND\_END\textgreater \\
\\
\#\# Example 4\\
Input:\\
user: can you check the trains on tuesday?\\
assistant: PLAN I'm going to look for trains on tuesday \textless COMMAND\_END\textgreater  APICALL \{"name": "search\_train", "parameters": \{"leaveAt": "", "destination": "", "day": "tuesday", "arriveBy": "", "departure": ""\}\} \textless COMMAND\_END\textgreater \\
user: APIRETURN []\\
Output:\\
PLAN It seems there are no trains leaving tuesday, I will tell the customer. \textless COMMAND\_END\textgreater  SPEAK It seems there are no trains leaving tuesday, does any other day work for you? \textless COMMAND\_END\textgreater \\
\\
\#\# Example 5\\
Input:\\
user: so I want a train leaving at 8am\\
assistant: PLAN I will find a train leaving at 8am by using the search\_train api call \textless COMMAND\_END\textgreater  APICALL \{"name": "search\_train", "parameters": \{"leaveAt": "08:00", "destination": "", "day": "", "arriveBy": "", "departure": ""\}\} \textless COMMAND\_END\textgreater \\
user: APIRETURN []\\
assistant: PLAN There are no trains that leave at 8am, I will tell the customer. \textless COMMAND\_END\textgreater  SPEAK Sorry, but there are no trains leaving at 8am. Are there any other times you'd like to look for? \textless COMMAND\_END\textgreater \\
user: Sure how about 9am?\\
Output:\\
PLAN I will look up a train using the search\_hotel api. The parameters will be parking and high pricerance. \textless COMMAND\_END\textgreater  APICALL \{"name": "search\_train", "parameters": \{"leaveAt": "09:00", "destination": "", "day": "", "arriveBy": "", "departure": ""\}\} \textless COMMAND\_END\textgreater \\
\\
\#\# Example 6\\
Input:\\
user: so I want a train leaving at 8am with 4 stars\\
assistant: PLAN I will find a train leaving at 8am with 4 stars by using the search\_train api call \textless COMMAND\_END\textgreater  APICALL \{"name": "search\_train", "parameters": \{"leaveAt": "08:00", "destination": "", "day": "", "arriveBy": "", "departure": "", "stars": "4"\}\} \textless COMMAND\_END\textgreater \\
user: APIRETURN ERROR\\
Output:\\
PLAN The api failed, I will double check the api definition by writing it out. definition: \{"name": "search\_train", "parameters": \{"leaveAt": "", "destination": "", "day": "", "arriveBy": "", "departure":""\}. I used a "stars" parameter which is invalid. Let me redo the api with the correct parameters \textless COMMAND\_END\textgreater  APICALL \{"name": "search\_train", "parameters": \{"leaveAt": "08:00", "destination": "", "day": "", "arriveBy": "", "departure": ""\}\} \textless COMMAND\_END\textgreater \\
\\
\#\# Example 7\\
Input:\\
user: I want a restaurant in the north\\
assistant: PLAN I will search for a restaurant in the north area using the search\_restaurant api call \textless COMMAND\_END\textgreater  APICALL \{"name": "search\_restaurant", "parameters": \{"food": "", "pricerange": "", "name": "", "area": "north"\}\} \textless COMMAND\_END\textgreater \\
user: APIRETURN [\{...\}, \{...\}, ...]\\
assistant: PLAN There are multiple options returned, I will inform the customer and ask them if theres any other constraints they are looking for. \textless COMMAND\_END\textgreater  SPEAK Ok, I found a lot of restaurants in the north, is there anything else you want? \textless COMMAND\_END\textgreater \\
user: Yes I also want it to have room service\\
Output:\\
PLAN I will look up a hotel using the search\_restaurant api. The parameters are north area and room service. \textless COMMAND\_END\textgreater  APICALL \{"name": "search\_restaurant", "parameters": \{"food": "", "pricerange": "", "name": "", "area": "north"\}\} \textless COMMAND\_END\textgreater \\
\\
\#\# Example 8\\
Input:\\
user: Wow what a pretty day!\\
Output:\\
PLAN The user commented on what a pretty day it is. I will reply that I agree. \textless COMMAND\_END\textgreater  SPEAK You're right, it is so pretty! \textless COMMAND\_END\textgreater \\
\\
\#\# Final Output\\
Input:\\
\bottomrule
\end{tabular}
\end{table*}

\begin{table*}[t]
\centering
\scriptsize
\caption{Goal based user simulator prompt}
\label{tab:goal-prompt}
\begin{tabular}{p{0.95\textwidth}}
\toprule
\textbf{SYSTEM:}\\
You're a customer talking to a travel agent. 
\\
You have the following goals you want to accomplish in the conversation (don't relay them all at once to the agent):
\\
\{goals\}
\\
\\
Discuss with the agent to try and accomplish each one of your goals in order. 
\\
If the agent fails at an action, check other goals for a backup plan
\\
Relay information piecemeal to the agent to encourage conversation.
\\
EXCEPTION: Make sure you've communicated all the neccessary information for that intent before proceeding with a booking.
\\
ALL of your listed goals must be fufilled in order for you to agree to a booking.
\\
DO NOT say <span class='emphasis'> or </span> to the agent.
\\
When you want to end the conversation say END\_CONVERSATION
\\
Always say END\_CONVERSATION to hang up!
\\
\bottomrule
\textbf{USER:}\\
REMEMBER: You are a customer talking to a travel agent.
\\
When you want to end the conversation say END\_CONVERSATION
\\
Always say END\_CONVERSATION to hang up!
\\
Try to address your next goal or finish the current goal you're focusing on.
\\
Note: if you are looking for a "place to stay", don't refer to it as a hotel unless the goals explicitly state you are looking for a type <span class='emphasis'>hotel</span>.
\\
Don't relay all the information about your goal to the agent at once.
\\
ABOVE ALL ELSE, it is critical ALL of your listed goals are fufilled in order for you to agree to a booking. Double check each of your requirements and tell the agent if one is not met. If you're not sure, double check.
\\
EXCEPTION: Make sure you've communicated all the neccessary information for that intent before proceeding with a booking.
\\
If the agent fails at an action, check other goals for a backup plan. 
\\
Remeber, you are the customer. 
\\
CUSTOMER:
\\
\bottomrule
\end{tabular}
\end{table*}

\begin{table*}[t]
\centering
\scriptsize
\caption{Guide based user simulator prompt}
\label{tab:guide-prompt}
\begin{tabular}{p{0.95\textwidth}}
\toprule
\textbf{USER:}\\
You are a coach giving tips to a user simulator trying to replicate a conversation as consistently as possible. The user simulator is in the middle of a conversation, give it advice on what to do in the next turn.
\\
Consistency means that over multiple runs, the user simulator should behave in the exact same way, it is your job to try and help it stay on the same trajectory every run.
\\
\#\#\#\#\#\# Grounding Goals and Conversation \#\#\#\#\#\#\#\#\#
\\
Customer goals:
\\
{goals}
\\
The following is the source conversation the user simulator is trying to replicate:
\\
\{goal\_convo\}
\\
\#\#\#\#\#\#\#\#\#\#\#\#\#\#\#\#\#\#\#\#\#\#\#\#\#\#\#\#\#\#\#\#\#\#\#\#\#\#\#\#\#\#\#\#\#\#\#\#\#\#\#
\\
\#\#\#\#\#\#\#\# CURRENT (real) Conversation \#\#\#\#\#\#\#\#\#\#\#\#\#\#\#\#\#\#\#\#\#\#\#
\\
This is the CURRENT conversaiton the user simulator is having:
\\
\{current\_convo\}
\\
Use your best judgement if the conversation is not going well, it's possible the agent is not good enough and you need to end the conversation. End the conversation by putting END\_CONVERSATION after your quote.
\\
Keep in mind the Customer goals all must be communicated in order to give the agent enough information to properly search and book.
\\
It is critical you give consistent advice over multiple iterations of the same conversation. The best way to do that is to ground your response in the source conversation and providing quotes whenever possible.
\\
Please write breif advice on what the user simulator should say in order to keep it consistent and aligned with the source conversation. Write this advice to the user simulatior, referring to it as "you". No yapping.:
\\
Example: 
\\
Advice:
\\
The user should ...
\\
Suggested quote:
\\
"Hello, how can I help you?"
\\
Advice:
\\
The conversation should be ended
\\
Suggested quote:
\\
"Thanks, goodbye" END\_CONVERSATION
\\
Output:
\\
\\
\bottomrule
\end{tabular}
\end{table*}